\documentclass{IET-Conf-Paper}

\usepackage{fancyhdr}

\begin{document}

\title{ARCHITECTURAL BLUEPRINT FOR HETEROGENEITY-RESILIENT FEDERATED LEARNING}

\author{Satwat Bashir, Tasos Dagiuklas, Kasra Kassai, Muddesar Iqbal}

\address{\add{1}{Department of Computer Science, London South Bank University, London, UK}
\email{\{bashis11, tdagiuklas, kasra.kassai, m.iqbal\}@lsbu.ac.uk}}


\begin{abstract}
This paper proposes a novel three-tier architecture for federated learning to optimize edge computing environments. The proposed architecture addresses the challenges associated with client data heterogeneity and computational constraints. It introduces a scalable, privacy-preserving framework that enhances the efficiency of distributed machine learning. Through experimentation, the paper demonstrates the architecture's capability to manage non-IID data sets more effectively than traditional federated learning models. Additionally, the paper highlights the potential of this innovative approach to significantly improve model accuracy, reduce communication overhead, and facilitate broader adoption of federated learning technologies. 
\end{abstract}

\maketitle
\thispagestyle{fancy}

\section{Introduction}

In the digital era, the proliferation of data from the Internet of Things (IoT) and smart devices presents unprecedented opportunities and challenges. Edge computing has become an essential paradigm, processing data proximal to its origin, thereby reducing latency and bolstering privacy. Nonetheless, the integration of Federated Learning (FL) with edge computing encounters significant obstacles, such as high communication costs and data/model heterogeneity, especially with non-Independently and Identically Distributed (non-IID) data \cite{Tang2024,Zhou2024, ResearchChallengesheterogeneous}. 

FL emerges as a revolutionary method, processing data locally on devices while sharing model updates instead of raw data, thus significantly diminishing bandwidth demands and fortifying privacy. Yet, FL confronts notable challenges, including substantial communication costs and the complexity of data/model heterogeneity across disparate devices \cite{Hasan2024}. The data collected from various devices often exhibit non-IID characteristics, meaning that the data points are neither independent nor identically distributed across the network. This variation poses significant challenges for traditional learning models, using edge-cloud computing, where data are processed close to its source \cite{Zhang2024, Consul2024}. Therefore, underscores the need for innovative solutions tailored to the demands of edge computing environments \cite{Sharma2024} . 

To address the nuanced challenges presented by edge computing and non-IID data, a novel three-layered architecture for FL has been introduced. This design seamlessly integrates \textit{ \textbf{clients}},  \textit{\textbf{edge layers}}, and \textit{\textbf{fedge layer}} to enhance data processing efficiency and model efficacy. By defining clear roles and interactions among the layers, the proposed architecture provides a focused strategy for addressing data and computational heterogeneity. The architecture introduces aggregation across two layers and multi-global models, allowing for the management of distinct models and streamlining aggregation to better address the concerns of data diversity \cite{Yang2024, Li2024}. 

This paper is structured as follows: Section 2 provides a background on FL, discussing the multifaceted nature of heterogeneity and its challenges. Section 3 presents the theoretical considerations for improved FL frameworks, introducing our novel three-tier architecture. Section 4 delves into empirical findings on FL with non-IID data, showcasing the architecture's performance through various scenarios. Section 5 concludes the study with a discussion on the implications of our findings and outlines future directions for research. Throughout, we aim to highlight the innovative aspects of our approach and its potential impact on the field of FL in edge computing environments.

\section{Background on Federated Learning}
The advent of FL represents a monumental shift towards decentralized model training across a myriad of devices, pivotal in the era dominated by IoT. This approach allows for the harnessing of vast amounts of data directly at the source, thus preserving user privacy and leveraging computational capabilities without the need for data centralization \cite{FL-BreastCancer}. Despite its potential, the practical deployment of FL is fraught with challenges, predominantly due to data heterogeneity, model variance, and the diverse capabilities of devices, which collectively demand innovative solutions for effective global model development \cite{ResearchChallengesheterogeneous}. 

\subsection{The Multifaceted Nature of Heterogeneity}

In FL, heterogeneity manifests in several critical areas: data, model, and device, each introducing unique challenges to the learning process. Data heterogeneity, particularly with non-Independently and Identically Distributed (non-IID) data across devices, leads to significant discrepancies in local model accuracy and global model performance \cite{FLforECResearchProblems, Adapterfl}. This variance underscores the difficulty of creating models that generalize well across diverse data distributions, a fundamental goal of FL. Model heterogeneity, stem from the diverse computational capabilities of devices, influencing the choice of model architectures and thus impacts the coherence of aggregated global models \cite{huang2019patient}. Furthermore, device heterogeneity, is characterized by disparities in computational power, memory, and connectivity, directly affects FL’s efficiency and scalability \cite{briggs2020federated, ghosh2019robust}.

Adding to these complexities, are the challenges posed by communication heterogeneity, which impacts the efficiency and reliability of data transmission, crucial for collaborative learning in distributed environments. These multifaceted aspects of heterogeneity highlight the intricate challenges FL faces, necessitating comprehensive strategies that address not only the technical hurdles, but also the ethical considerations inherent in deploying FL in real-world scenarios \cite{mansour2020three, kim2021dynamic}. 

\subsection{Tackling Non-IID Data Challenges}

Non-IID data significantly impedes FL’s model performance and learning efficiency. To counteract this, innovative approaches such as Personalized Federated Learning (PFL) have been proposed, focusing on architecture-based and similarity-based personalization \cite{tan2022towards}. Techniques like parameter decoupling and knowledge distillation aim to tailor models to individual device capabilities while drawing insights from the global model \cite{stallmann2022towards, duan2023combining} . Despite these advancements, achieving equitable and efficient learning in federated settings remains a formidable challenge, highlighting the need for further innovation \cite{DisasterMTL}. 

\subsection{Critical Analysis of Aggregation Methods}

At the heart of FL lies the model aggregation process, with FedAvg being the prototypical method \cite{konevcny2016federated,mcmahan2017communication,banabilah2022federated}. While FedAvg has shown effectiveness under conditions of homogeneity, its performance significantly diminishes in the face of data heterogeneity, leading to a loss of precision in local models \cite{tan2022towards,shamsian2021personalized}. 

In the study referenced as \cite{pillutla2022federated}, the authors evaluate the impact of utilizing pre-trained models on local devices, focusing on the aggregation of shared parameters. They find that employing pre-trained models and sharing a subset of parameters from the start can enhance convergence speed, improve performance, and lower communication costs. Similarly, the work cited as \cite{achituve2021personalized} introduces pFedGP, which utilizes a shared kernel function to increase the accuracy of local models. However, this approach is constrained by the size of the dataset due to the limitations of the shared kernel function. To tackle the issue of communication overhead in personalized model training, the study \cite{ozkara2021quped}  applies soft quantization and transfer learning techniques, achieving higher accuracy for local devices and enhancing communication efficiency. Likewise, \cite{zhang2021parameterized} employs transfer learning to categorize devices with similar data and facilitate collaborative training for varied data sets.

Other research, such as \cite{liang2020think}, focuses on incorporating personalized layers through fine-tuning techniques to train the global model's shallow layers on each device. This method involves sharing only the foundational layers during the aggregation process, but it faces challenges related to storing the personalized model on each local device. Some studies, like \cite{t2020personalized}, address the issue of data heterogeneity by framing it as a regularization problem, utilizing Moreau Envelopes to improve the global model's convergence. However, a notable limitation of these approaches is their reliance on numerous adjustable parameters.

Despite these innovative strategies, limitations remain, particularly in handling complex real-world data distributions, ensuring communication efficiency, and accommodating diverse scenarios. The pursuit of more adaptable and robust solutions continues, reflecting the evolving landscape of FL research \cite{xia2021survey,kairouz2021advances}. 

\subsection{Towards Comprehensive Solutions}

The exploration of aggregation methods and personalization strategies within FL reveals a concerted effort to address the challenges posed by heterogeneity and non-IID data as demonstrated in studies \cite{DisasterMTL, FL-BreastCancer}. However, the limitations of current approaches underscore the pressing need for adaptable, sophisticated solutions capable of navigating the intricacies of FL environments. The shift towards advanced edge-cloud models, the employment of cutting-edge algorithms, and the development of multi-model systems illustrate the ongoing quest for enhanced FL frameworks \cite{evolution}. Our findings lay the groundwork for a transformative FL ecosystem, characterized by its robustness, efficiency, and unparalleled scalability across myriad applications. Looking ahead, we explore advanced machine learning algorithms that could further refine the proposed architecture's ability to manage extreme data heterogeneity. Additionally, we aim to investigate the implications of our framework in real-world scenarios, such as IoT and mobile computing, where FL's potential can be fully unleashed. 

\section{Theoretical Considerations for Improved FL
Frameworks}

In the rapidly evolving field of FL, identifying architectures capable of handling the complexities of distributed data processing is critical. Traditional FL models have pioneered the way forward, but often grapple with limitations due to non-IID data distributions and significant computational demands on client devices. In response to these challenges, this paper proposes a novel three-tier FL architecture designed to enhance data privacy, computational efficiency, and scalability across distributed networks. 

\subsection{Rationale for a Multi-Global Model Framework}

\begin{figure}[h]
    \centering
    \includegraphics[width=1\linewidth]{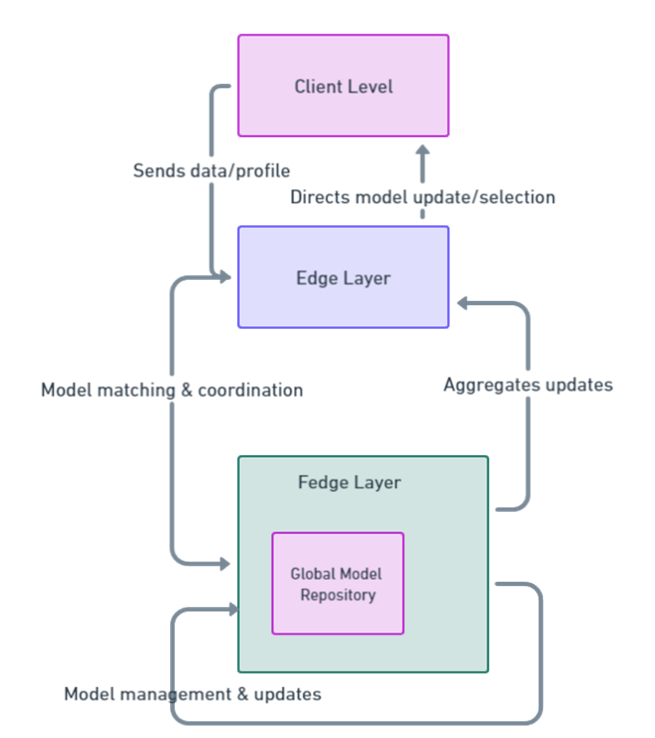}
    \caption{Proposed three-layered architecture, delineating the interactions between the client level, edge layer, and fedge layer}
    \label{architecture flow diagram}
\end{figure}

The proposed architecture significantly diverges from traditional FL paradigms by implementing a multi-global model strategy within a hierarchical framework. This approach is analyzed for its potential to facilitate personalized and efficient learning across heterogeneous devices and data distributions. The operational mechanics of each layer are shown in Figure  \ref{architecture flow diagram}, illustrating how they collectively contribute to model convergence and scalability. Preliminary simulations underscore the adaptability and performance of our architecture when compared to standard FL frameworks, particularly under non-IID conditions. 

\textbf{The Client Layer}, profiling ensures that each participant begins with a model specifically tailored to its unique data set. This enables effective local training, optimizing computational load and aligning model evolution with individual data characteristics. Empirical evidence demonstrates that this alignment results in improved initial accuracy and faster convergence rates.

\textbf{The Edge Layer} acts as a critical coordinator, analyzing metadata from client training sessions to assess model states and requirements. If a suitable model match for a client is not found within its repository, the edge layer queries the fedge layer. This process ensures that clients are always provided with the most appropriate model for their data, enhancing the learning process's efficiency and personalization.

\textbf{The Fedge Layer} is responsible for storing multiple distinct global models, managing them to ensure they are up-to-date and accurately reflect the diverse data landscapes they represent. When requested by the edge layer, the fedge Layer searches for a suitable model match. If no match is found, it informs the edge layer, prompting the creation of a new model for the client. This new model is then saved as a global model, enriching the fedge Layer's repository. Furthermore, the fedge Layer periodically updates its global models based on inputs from the edge layer, performing aggregation and updates of similar models to maintain a current and effective suite of global models.

This structured approach tackles computing issues and helps create better learning environments. We have carefully followed principles of technical precision in our work, making sure our approach, experiments, and results are shared accurately and clearly. 

\subsection{Innovation}

The core innovation of the proposed architecture is its deployment of multiple global models, a strategy unprecedented in current FL frameworks. This supports personalized learning for diverse client datasets and introduces a novel method of model sharing and aggregation that significantly reduces computational overhead. The empirical findings affirm this architecture's potential to redefine FL paradigms, offering a scalable solution for real-time data analysis.

\section{Empirical Findings on Federated Learning with Non-IID Data}
This section delves into the performance of FL, specifically focusing on the Federated Averaging (FedAvg) method, under non-IID data conditions. This investigation is supported by a  experimental setup, revealing insights into FL's adaptability to varied data distributions. 

\subsection{System Architecture and Experimental Setup}

\textbf{Dataset:} The analysis employs the MNIST dataset, a benchmark for evaluating machine learning models through digit recognition tasks, featuring 70,000 grayscale images across ten classes \cite{hojjatk_mnist}. 
\subsection{Scenario Design}
We aim to explore the multifaceted nature of non-IID distributions in a controlled yet challenging environment. For this, we designed two distinct non-IID scenarios using the MNIST dataset to test the FL architecture's handling of data heterogeneity: By simplifying the scenario to these two configurations, we not only highlight the limitations of traditional FL model in handling such disparities, but also establish a clear, proof-of-concept framework for demonstrating the superior adaptability and performance of our proposed three-layer architecture. This approach allows to systematically assess the architecture's efficacy in managing non-IID data, ensuring that the findings have practical implications for enhancing FL in heterogeneous environments.

\begin{itemize}
    \item \textbf{Generalizable Non-IID Scenario}: The MNIST dataset has been divided among three clients to simulate a non-IID but generalized scenario. An equal number of images has been allocated to each client. Such allocation provides restriction to distinct labels, ensuring diversity in data distribution while maintaining a level of generality across the dataset.
    \item \textbf{Non-Generalizable Non-IID Scenario}: For a more challenging setup, we have created another non-IID scenario by dividing the MNIST dataset between two clients. One client has received images from only two labels with a robust dataset size for training, while in the second client has been allocated images from the remaining eight labels, presenting  a challenging test for the architecture's efficiency. 
\end{itemize}
 
\textbf{Client-Server and Three-Tier FL Models:} We compare the traditional client-server FL model with our proposed three-tier FL model, highlighting the latter's enhanced data privacy, computational efficiency, and scalability.

\subsection{\textbf{Methodology and Model Implementation} } 
\begin{figure}[h]
    \centering
    \includegraphics[width=1\linewidth]{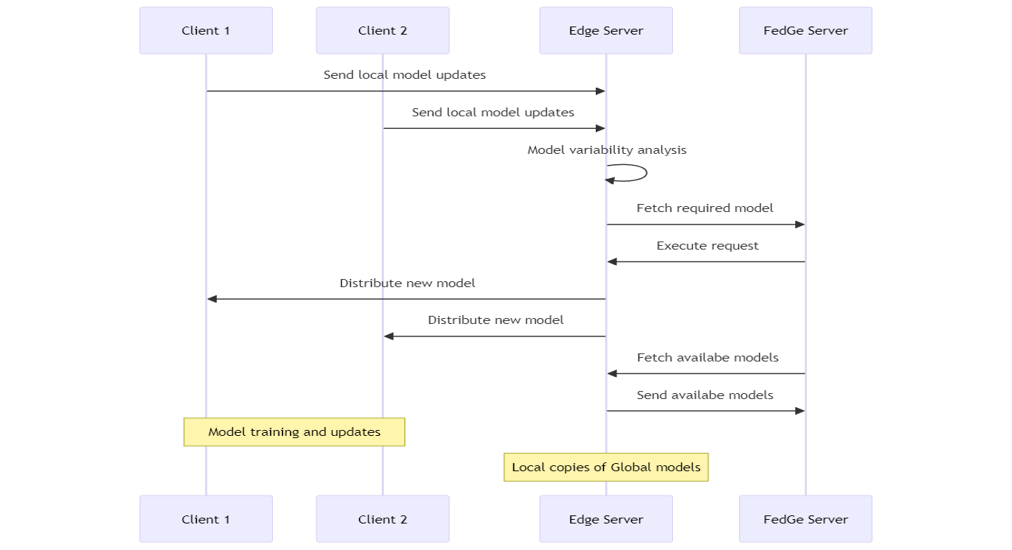}
    \caption{The sequence diagram illustrates the dynamic interactions within the proposed architecture, focusing on the multi-global model strategy within a hierarchical framework }
    \label{sequence diagram}
\end{figure}

\textbf{FL Algorithms and Model Details:} The FedAvg algorithm benchmarks the standard FL approach, while the proposed three-tier FL model introduces a specialized algorithm to better manage non-IID data through a multi-global model framework. The dynamic interactions and the structural foundation of the proposed three-tier FL model is shown in Figure \ref{sequence diagram}. This sequence diagram elucidates the multi-global model strategy implemented within a hierarchical framework, showcasing the model's innovative approach to managing non-IID data. 

\textbf{Model Specification:} A Simple Convolutional Neural Network (SimpleCNN) has been chosen for consistency in evaluating both FL models against the MNIST dataset.

\subsection{Parameters and Data Handling} 
\textbf{Parameters and Configurations:} The experimental setup involves multiple communication rounds, a learning rate of 0.01, momentum of 0.5, and a batch size of 64.

\textbf{Data Preprocessing:} Standard normalization processes has been applied to ensure uniform data scaling across clients.

\textbf{Model Serialization:} We detail the serialization process for both FL models, emphasizing the efficient transmission of model updates between clients, edge layers, and the global model repository.

\subsection{Findings, Behaviors, and Insights}

\begin{figure}[h]
    \centering
    \includegraphics[width=1\linewidth]{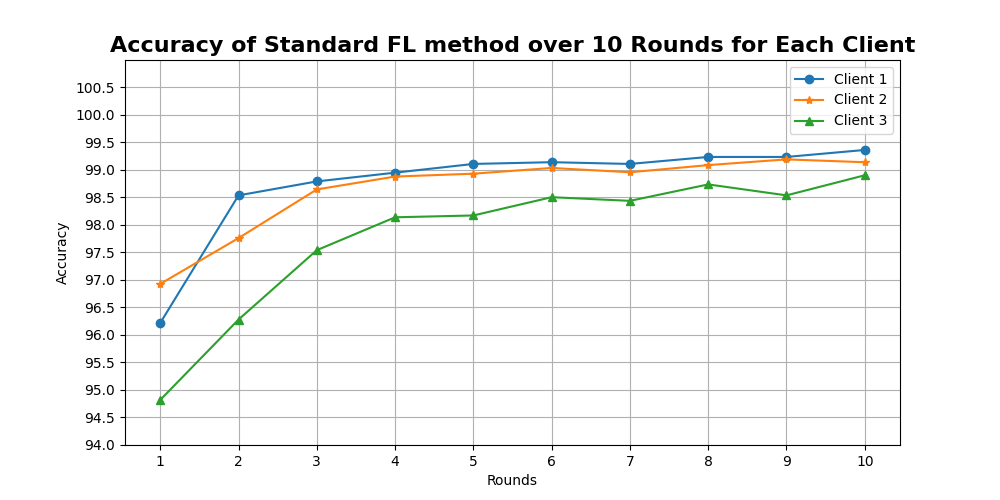}
    \caption{Accuracy of Standard FL for non-IID Scenario 1}
    \label{fig:SFL-S1}
\end{figure}

In the first scenario, which tested the models' response to label skew with a balanced data distribution, the standard FL model showed a progressive increase in accuracy. Client 1's accuracy began at 96.22\% and rose to 99.36\%. For Client 2, the accuracy started at 96.93\% and reached 99.14\%, and Client 3 improved from 94.82\% to 98.90\%. These trends are depicted in Figure \ref{fig:SFL-S1}.

\begin{figure}[h]
    \centering
    \includegraphics[width=1\linewidth]{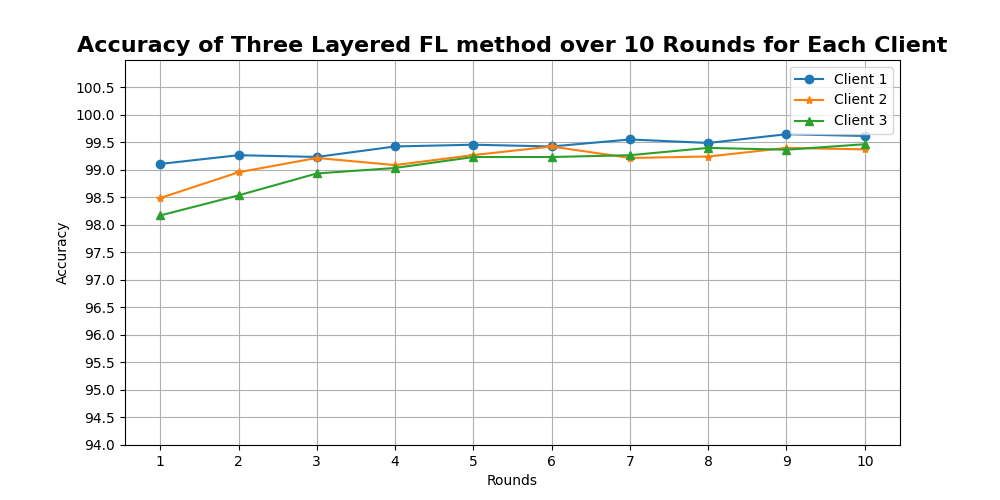}
    \caption{Accuracy of Three Layered FL for non-IID Scenario 1}
    \label{fig:3FL-S1}
\end{figure}

The three-tier FL model demonstrated a consistent high accuracy across all clients from the start, with Client 1 reaching 99.62\%, Client 2 ending at 99.38\%, and Client 3 at 99.47\% by the tenth round. These findings indicate the three-tier model's ability to quickly achieve and maintain high accuracy, as shown in Figure \ref{fig:3FL-S1}.

\begin{figure}[h]
    \centering
    \includegraphics[width=1\linewidth]{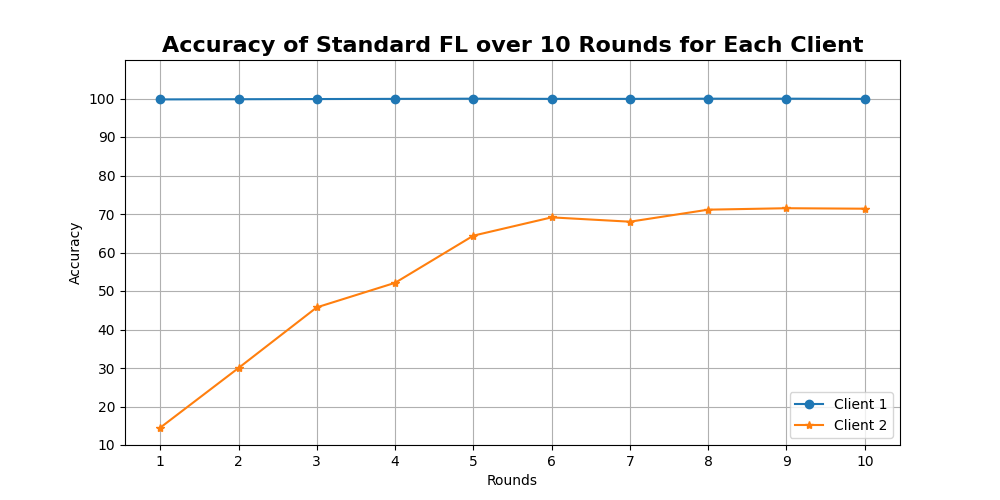}
    \caption{Accuracy of Standard FL for non-IID Scenario 2}
    \label{fig:SFL-S2}
\end{figure}

In the second scenario, designed to present a more complex non-IID situation with both label and data skew, the standard FL model and the three-tier FL model were subjected to a challenging test. For the standard FL model, Client 1 managed to maintain an accuracy of nearly 99.86\%, while Client 2, with less representative data and a broader label distribution, started at a lower 14.5\% accuracy, gradually increasing to 71.38\% as shown in Figure \ref{fig:SFL-S2}. This growth trajectory signifies the model's gradual adaptation to the non-IID conditions.

\begin{figure}[h]
    \centering
    \includegraphics[width=1\linewidth]{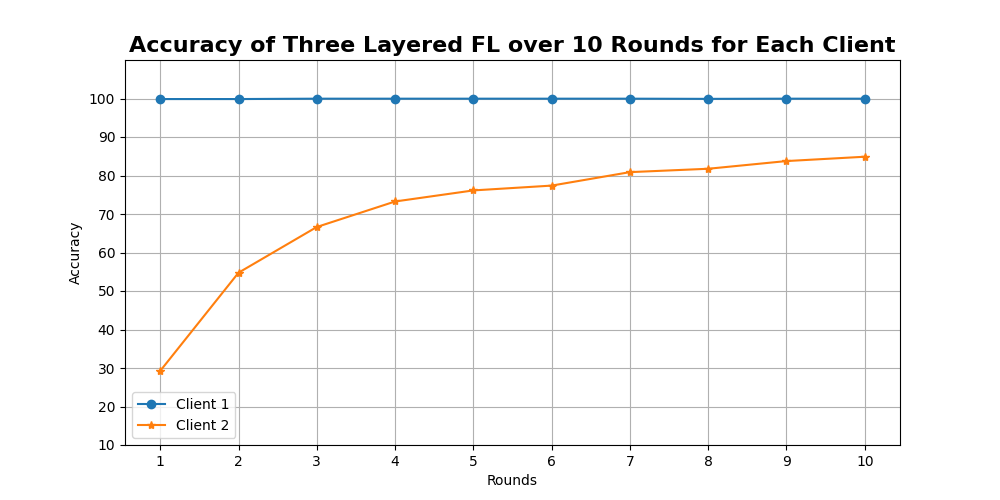}
    \caption{Accuracy of Three Layered FL for non-IID Scenario 2}
    \label{fig:3FL-S2}
\end{figure}

Conversely, the three-tier FL model excelled, with Client 1 maintaining a near-perfect accuracy of approximately 99.95\%, and Client 2, despite the sparse data representation and more extensive label range, exhibited a substantial improvement, escalating from 29.25\% to 84.88\% accuracy. This remarkable progress is graphically represented in Figure \ref{fig:3FL-S2}.

These results validate the standard FL model's capacity to achieve generalization over time in balanced scenarios, but they also spotlight the three-tier FL model's advanced capability to handle various degrees of non-IID data efficiently. The superior performance of the three-tier model, especially for Client 2 in the second scenario, underlines its effectiveness in real-world applications, where data distributions and label availability can significantly differ. The model's robustness in enhancing learning under such complex conditions attests to its potential for broad deployment and underscores the value of the proposed FL technique. 

\section{Conclusion}

This study introduced a novel three-layered architecture for FL, designed to address the significant challenges of integrating FL with edge computing, especially under conditions of non-IID data. Our architecture demonstrates a promising potential to enhance FL's applicability in edge computing environments by optimizing training and processing efficiencies. By introducing an intermediary layer, the fedge layer, and employing a multi-global model framework, we provided a focused strategy for managing client and computational heterogeneity. Empirical evaluations, utilizing the MNIST dataset under various non-IID scenarios, have shown that our proposed architecture outperforms traditional FL frameworks in terms of accuracy, scalability, and efficiency.

However, it's important to acknowledge the limitations of our current work. The proposed architecture, while innovative, serves as a foundational proof of concept. It is important to acknowledge that the architecture's effectiveness, as demonstrated under controlled experimental conditions, may vary in complex real-world scenarios. The adaptation to different types of non-IID data, the management of communication overhead, and the practical deployment in edge environments with varied device capabilities and network conditions are aspects that require further exploration and validation.

In conclusion, this paper contributes a significant step forward in the pursuit of more efficient, privacy-preserving, and scalable FL solutions tailored for edge computing. The proposed three-layered architecture, with its emphasis on addressing non-IID data challenges and client heterogeneity, lays the groundwork for future advancements in the field. By continuing to refine and expand upon this framework, we can move closer to realizing the full potential of FL in the edge computing paradigm.

 \section{Future Directions}

In future work, we plan to further refine and validate the effectiveness of our proposed three-layer architecture by testing it alongside the latest and most innovative federated learning approaches. This comparative analysis will provide valuable insights into where our model stands in the current research landscape and identify areas for improvement. Additionally, experimenting with other benchmark datasets beyond MNIST will allow us to assess the versatility and robustness of our architecture across different types of data and learning scenarios. This step is crucial for ensuring that our model can be effectively applied in various domains and with diverse data characteristics.

Furthermore, we aim to enhance the functionalities and operation of the fedge layer to better suit real-world scenarios. This includes optimizing the fedge layer for more efficient model management and updating processes, which are vital for handling the dynamic and diverse nature of data in practical applications. Another critical area of development will be the refinement of the client selection method. By designing more sophisticated criteria for client selection, we can ensure that our architecture is adaptable to real and broader contexts, improving its applicability and efficiency in distributed learning environments.

By focusing on these areas, we anticipate not only strengthening the foundation of our proposed architecture but also significantly advancing its potential for practical implementation in federated learning systems.

\section*{References}

\bibliographystyle{IEEEtran}

\bibliography{IET}

\end{document}